\documentclass[letterpaper, 10 pt, conference]{ieeeconf}  % Comment this line out
\IEEEoverridecommandlockouts                              % This command is only
\usepackage{xcolor}
\usepackage{comment}
\usepackage{graphics} % for pdf, bitmapped graphics files
\usepackage{epsfig} % for postscript graphics files
\usepackage{amsmath} % assumes amsmath package installed
\usepackage{amssymb}  % assumes amsmath package installed
\usepackage{mathtools}
\usepackage{marginnote}
\usepackage{hyperref}
\usepackage{cite}
\usepackage{mathrsfs}
\usepackage{algorithm,algorithmic}
\usepackage{xcolor}
\usepackage[inkscapelatex=false]{svg}
\usepackage{multirow} % Add this to the preamble
\usepackage{array} 
\usepackage{etoolbox}
\usepackage[thicklines]{cancel}
\usepackage[inkscapelatex=false]{svg}
\usepackage{flushend}

\makeatletter
\patchcmd{\@makecaption}
  {\scshape}
  {}
  {}
  {}
\makeatletter
\patchcmd{\@makecaption}
  {\\}
  {.\ }
  {}
  {}
\makeatother
\def\tablename{Table}

\title{\LARGE \bf
Simultaneous Ground Reaction Force and State Estimation via Constrained Moving Horizon Estimation} % may changeA the name 

\author{ Jiarong Kang and Xiaobin Xiong   % stops a space
\thanks{The authors are with the Wisconsin Expeditious LeggedAI Lab (WELL-Lab) at the University of Wisconsin-Madison.
        {Corresponding to \tt\small xiaobin.xiong@wisc.edu}.}%
}

\begin{document}

\newcommand{\Kang}[1]{{\color{blue} #1}}

\newcommand{\TODO}[1]{{\color{blue} \textbf{TODO: #1}}}

\newcommand{\block}[1]{\noindent{\textbf{#1}:}}
\newcommand{\emphhh}[1]{{\color{yellow} \textbf{#1}}}

\maketitle
\thispagestyle{empty}
\pagestyle{empty}

%%%%%%%%%%%%%%%%%%%%%%%%%%%%%%%%%%%%%%%%%%%%%%%%%%%%%%%%%%%%%%%%%%%%%%%%%%%%%%%%
\begin{abstract}
Accurate ground reaction force (GRF) estimation can significantly improve the adaptability of legged robots in various real-world applications. For instance, with estimated GRF and contact kinematics, the locomotion control and planning assist the robot in overcoming uncertain terrains. The canonical momentum-based methods, formulated as nonlinear observers, do not fully address the noisy measurements and the dependence between floating-base states and the generalized momentum dynamics. In this paper, we present a simultaneous ground reaction force and state estimation framework for legged robots, which systematically addresses the sensor noise and the coupling between states and dynamics. With the floating base orientation estimated separately, a decentralized Moving Horizon Estimation (MHE) method is implemented to fuse the robot dynamics, proprioceptive sensors, exteroceptive sensors, and deterministic contact complementarity constraints in a convex windowed optimization. The proposed method is shown to be capable of providing accurate GRF and state estimation on several legged robots, including the custom-designed humanoid robot Bucky, the open-source educational planar bipedal robot STRIDE, and the quadrupedal robot Unitree Go1, with a frequency of 200Hz and a past time window of 0.04s.
\end{abstract}

\section{Introduction}

Legged robots have demonstrated exceptional versatility, with applications spanning various fields \cite{Anymal}, including logistics, search and rescue, inspection, agriculture, and entertainment. Their ability to traverse complex terrains and environments makes them indispensable in scenarios where wheeled or tracked robots face limitations. For reliable performance in dynamic settings, legged robots require a combination of accurate estimation and robust control strategies. These components are critical for maintaining balance, ensuring stability, and enabling autonomous behaviors.

Recent advancements in legged robot control have demonstrated that force control can effectively enable dynamic locomotion across diverse terrains and environments. This approach utilizes planned ground reaction force (GRF) as control inputs, ensuring stability and adaptability in varying conditions. As a result, an accurate estimation of both GRF and robot states can greatly enable the control capability of force-based controllers \cite{Stance}, and enhance the perception capabilities of the environment on legged robots~\cite{soft_est}. 

Installing contact sensors, such as pressure sensors \cite{UNTREEWebsite}, on legged robots is a common and straightforward method for providing a binary indicator of contact states. However, only a few robots are equipped with multi-axis force/torque (F/T) sensors \cite{Xijilefu_QP, flying_baby}, due to their high cost, weight, and fragility. GRF estimation using a general sensor setup, i.e., IMUs, joint encoders, joint current sensors, stereo cameras, and foot contact sensors, can offer a more practical, efficient, and effective solution.

% To overcome the limitations of sensor selection, current methods for estimating GRF rely on the Momentum-Based Observer (MBO) \cite{DBO_arm} considering the generalized momentum dynamics of the robot. While originally introduced for manipulators, the method has been applied in quadrupedal robots \cite{Contact_fusion}, \cite{View}, and humanoid robots \cite{humanoid_MBO}. However, compared with robot manipulator, legged robots are typically described by full-body dynamics, where the floating base states form part of the generalized coordinates used to characterize their motion.

To overcome sensor limitations, current methods for estimating the GRFs often rely on the Momentum-Based Observer (MBO) \cite{DBO_arm}, which incorporates the generalized momentum dynamics of the robot. Originally developed for manipulators, this method has demonstrated its effectiveness on quadrupedal robots \cite{Contact_fusion}, \cite{View}, and humanoid robots \cite{humanoid_MBO}. \cite{DisturbKF} further enhanced this approach by introducing a momentum-based Disturbance Kalman Filter (DKF) to stochastically address sensor noises and model uncertainties.

However, unlike manipulators, legged robots have dynamics that depend on floating base states - a relationship that previous works have not fully addressed. \cite{Contact_fusion}, \cite{View}, and \cite{humanoid_MBO} estimated GRF and floating-base states separately, creating a disconnect that limits the ability to fully capture the system's dynamics. Furthermore, in scenarios involving rigid contact, contact complementarity constraints \cite{complementarity} have been shown to enhance estimation accuracy during sliding and collision events \cite{lcpconstrainedKF}. Therefore, an optimization-based estimator that simultaneously estimates GRF and states, while incorporating constraint information, offers an ideal solution.
\begin{figure}[t]
    \centering
    \setlength{\abovecaptionskip}{0pt}
    \includegraphics[width=.99\linewidth]{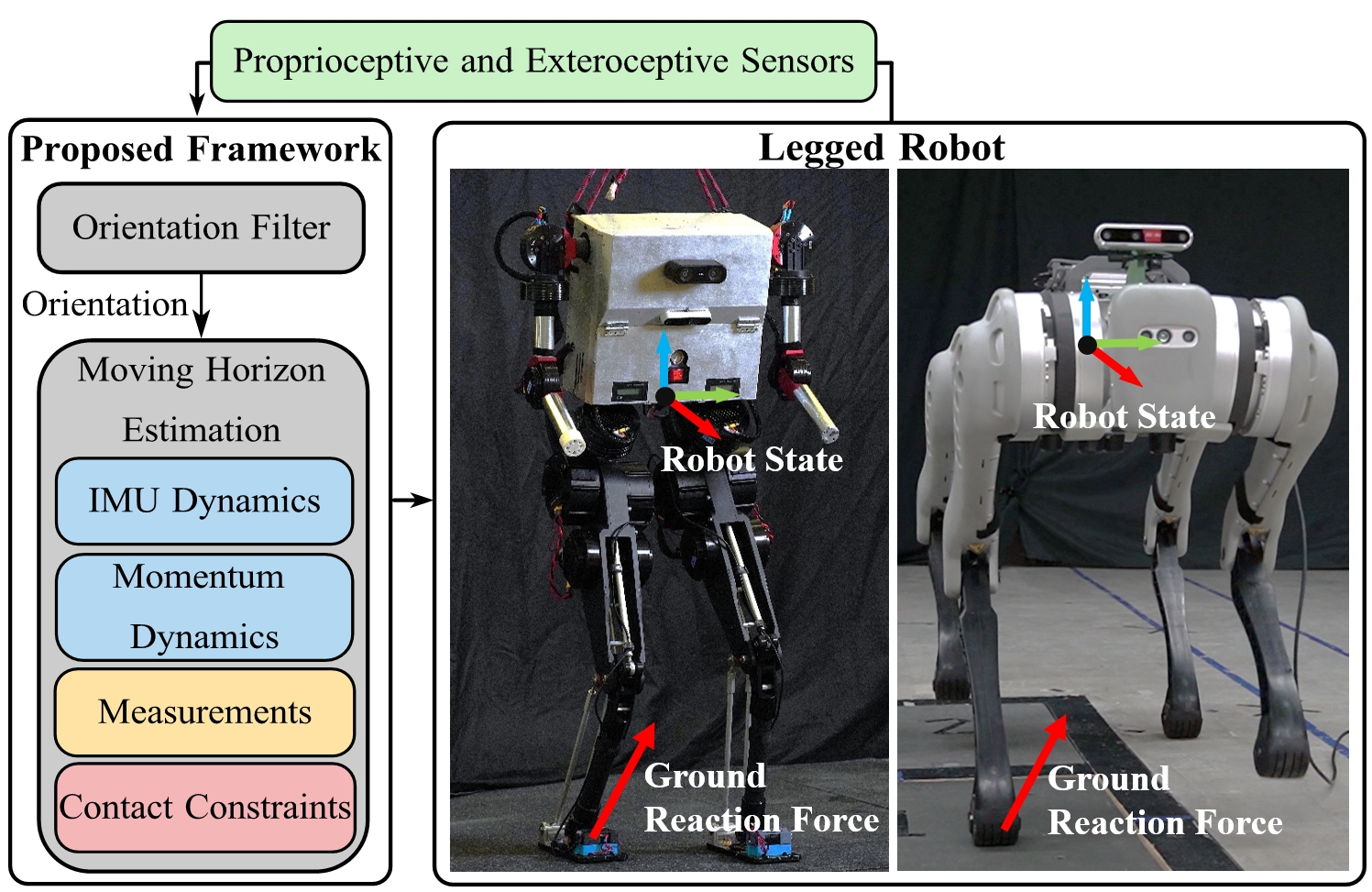}
    % \includesvg[width=1.0\linewidth]{figure/Framework.svg}
    \vspace{-10pt}
    \caption{Illustration of the proposed decentralized estimation framework for estimation on various legged robots, including the humanoid robot Bucky (left) and the quadrupedal robot B1 (right). The experiment video is available at: \href{https://youtu.be/Bih7cslSkTo}{\texttt{https://youtu.be/Bih7cslSkTo}}.}
    \label{Framework}
    \vspace{-20pt}
\end{figure}
 % The momentum dynamics depends on the estimate of the floating base states, and no constraints information can be included in the estimation. A simultaneous estimation of both the ground reaction force and floating base states, considering all on board sensors and physical constraints information will be the ideal setup.  

% In this paper, we aim to develop a contact-constrained MHE method that achieves simultaneous ground reaction force and state estimation. With the orientation estimated separately from a decentralized orientation filter as in \cite{kang24}, the nonlinear estimation problem is formulated as constrained Linear MHE and solved as Quadratic Program (QP).

In this study, we extend our previous decentralized MHE framework in \cite{kang24} to achieve simultaneous estimation of the GRF and floating-base states of the legged robot, and we demonstrate its superior performance compared to existing methods. The framework structure is illustrated in Fig. \ref{Framework}. With orientation estimated separately by another orientational filer or an Attitude and Heading Reference System (AHRS), the MHE is formulated and solved as a Quadratic Program (QP), which allows real-time estimation. This study makes the following contributions:
\begin{itemize}
    \item A novel state estimation framework based on decentralized MHE \cite{kang24}, enabling simultaneous GRF and state estimation by fusing proprioceptive and exteroceptive sensors with physical constraints in real time.
    \item Evaluation of the proposed method on the custom-designed humanoid robot Bucky, the open-sourced planar bipedal robot STRIDE \cite{Stride} and the commercial quadrupedal robot Unitree Go1 \cite{UNTREEWebsite}. 
    \item Comparison of the proposed method with conventional approaches, highlighting performance improvements from exteroceptive sensors and contact constraints.
\end{itemize}

\section{Related Work}
Existing research in the literature on external force estimation, particularly GRF estimation, has been approached in various ways. These methods can be generally classified into sensorized and sensorless approaches. 
% Sensorized methods typically rely on expensive multi-axis force/torque (F/T) sensors, which are often heavy, fragile, and increase the foot’s inertia, making them less suitable for dynamic legged locomotion. 
Several innovative sensorized feet have been developed recently for GRF estimation. For example, \cite{foot_vision} introduced a vision-based sensorized foot that is capable of estimating both GRFs and terrain information simultaneously. Nevertheless, the use of deformable foot mechanisms can complicate general control and estimation tasks. \cite{foot_imu} introduced a lightweight, robot-agnostic solution with an IMU mounted on the foot of the robot. However, it primarily focuses on contact detection rather than precise force estimation. 

To address the challenges associated with sensor design, sensorless methods have tackled the GRF estimation problem using a dynamics-based approach. In \cite{InvDyna_humanoid}, the disturbance caused by GRFs on the humanoid robot HRP-4 is directly obtained by comparing the full-body dynamics with measured or estimated accelerations of the generalized coordinate. The effectiveness of this method is highly dependent on the accuracy of the sensor data. Momentum-based methods mitigate this requirement by estimating the GRF using momentum dynamics. \cite{Contact_fusion}, \cite{View} estimated the GRF on the quadrupedal robot with a Momentum-Based Observer (MBO). However, these methods handle noise similarly to low-pass filters, without accurately addressing the sensor noise or bias. Additionally, despite that the dynamics and configuration of the robot depend on the floating-base states, these methods estimate the floating-base states and GRF separately using a loosely coupled approach, which is both redundant and sub-optimal.

Considering modeling uncertainties and sensor noise, \cite{DisturbKF} proposed a momentum-based Disturbance Kalman Filter (DKF) for force estimation on a robotic manipulator. To further address the correlation between floating-base estimation and force estimation, \cite{Xijilefu_QP} developed a constrained Quadratic Program (QP) based estimator that leverages both full-body dynamics and 3-axis F/T sensors on Atlas to estimate contact forces and robot states. However, the estimator performance is only partially demonstrated and potentially limited by the capabilities of the F/T sensor, since the results focus solely on GRF estimation in the vertical direction.

Recent research in \cite{DDP_MHE} formulated a nonlinear MHE to achieve a simultaneous estimation of forces, parameters, and states using holonomic-constrained full-body dynamics \cite{Constrained_dyn}. However, due to the computational demands of nonlinear optimization, the algorithm runs offline, and no estimated GRF results are provided. In contrast, our approach aims to achieve real-time simultaneous GRF and state estimation using decentralized MHE \cite{kang24}. This method integrates momentum dynamics, exteroceptive sensors, and contact complementarity constraints to enhance estimation performance.

\section{Preliminaries}
We first present the preliminaries of Moving Horizon Estimation~\cite{MHE}, orientation estimators, and the momentum dynamics models for the legged robot. 
\subsection{Moving Horizon Estimation Formulation}
A dynamic system is generally described by a nonlinear difference equation and a measurement equation:
\begin{equation}
    x^{+} = f(x,u) + \delta^x,\quad y = h(x) + \delta^y,\label{eq:dynamics}
\end{equation}
where $(\cdot)^+$ denotes the updated state after a fixed interval $\Delta t$, $f(\cdot)$ and $h(\cdot)$ denote the process model and measurement model, respectively. $x$, $u$, and $y$ denote the state, control input, and measurement of the system, respectively. $\delta^x \sim \mathcal{N}(0,Q)$ and $\delta^y \sim \mathcal{N}(0,R)$ represent the process noise and measurement noise subject to zero-mean Gaussian distributions.
Additionally, legged robots are often subject to various physical constraints, which can generally be expressed as state constraints $z(x) \leq 0 $.
% \begin{equation}
%     z(x) \leq 0 .\label{eq:physical_constraints}
% \end{equation}

The MHE solves a Maximum A Posteriori (MAP) problem for a trajectory of states $x_{[T-N:T]}$. It incorporates the control inputs  $u_{[T-N:T-1]}$ and measurements $y_{[T-N,T]}$ within a receding window of size $N+1$, spanning from time index $T-N$ to $T$. The notation $(\cdot)_{[i,j]}$ denotes the sequence of vectors starting from the time index $i$ and ending at the time index $j$. The solution at each time step is obtained by solving the MAP problem, subjects to the physical constraints:
\begin{align}
    \underset{x_{[T-N,T]} \in \mathcal{X} }{\text{argmax}}&\ {P(x_{[T-N,T]} |u_{[T-N,T-1]}, y_{[T-N,T]})} \textstyle \label{eq:MAP}, 
\end{align}
where $P$ denotes the posterior probability, and $\mathcal{X}$ represents the feasible state region considering the imposed constraints.
The MHE, formulated as an optimization problem, minimizes the negative log-likelihood of \eqref{eq:MAP} to compute the optimal estimation of $\mathbf{X}_{[T-N:T]}$:
\begin{equation}
 \mathbf{X}_{[T-N:T]} = \begin{bmatrix}
     x_{[T-N,T]} &
     \delta^x_{[T-N,T-1]} &    
     \delta^y_{[T-N,T]} \end{bmatrix}, \textstyle \label{eq:state}
\end{equation}
integrating the process and measurement model along with the physical constraints into the optimization:
\begin{align}
&\min_{\mathbf{X}_{[T-N,T]}} \  \Gamma(x_{T-N}) \textstyle \nonumber \\
&+ \textstyle{\sum}_{k = T-N}^{T-1} || \delta^x_{k} ||_{Q_k} + \textstyle{\sum}_{k=T-N}^T || \delta^y_k ||_{R_k} ,  \label{equ:MHE} \tag{MHE} \textstyle \nonumber\\
\text{s.t.}~~&y_k = h(x_k) + \delta^y_k,\ \forall k \in \{T-N,...,T \},\label{eq:measurement_constraints} \\
 x_{k+1}  &= f(x_k, u_k) + \delta^x_k,~\forall k \in \{T-N,...,T-1 \}, \label{eq:dynamic_constraints}\\
 &z(x_k) \leq 0, \ \forall k \in \{T-N,...,T \}, \label{eq:contact_constraints}
\end{align}
where $\Gamma(x_{T-N})$ denotes the arrival cost. This term enables the processing of a finite horizon of measurements by incorporating all information prior to the current horizon into the arrival cost. The standard approach computes the arrival cost using an Extended Kalman Filter \cite{MHE}, and recent research also proposes a probabilistic approach to derive this term from the Hessian matrix of \eqref{equ:MHE} \cite{probMHE}. In this paper, we adopt the arrival cost calculation method from our previous work \cite{kang24}, which involves solving the Schur complement of the KKT matrix for the equality-constrained MHE.

\subsection{Orientation Estimation Formulation}
Following prior practices in \cite{cheetah3}, \cite{HLIP}, the decentralized MHE \cite{kang24} utilizes a standalone and lightweight estimator for orientation estimation. Recently, off-the-shelf Attitude and Heading Reference Systems (AHRS) have demonstrated their ability to provide reliable orientation measurements $\hat{R}_{\mathcal{WB}}$, relative to gravity and the Earth's magnetic field \cite{madgwick}. For legged robots in static contact scenarios, filter-based state estimation methods, such as the Invariant-Extended Kalman Filter (InEKF) \cite{InEKF}, incorporate additional sensor data to deliver accurate orientation estimates. Generally, without a high-quality IMU sensor, vision information is the primary additional information needed to provide accurate orientation estimation. In our previous work \cite{kang24}, a vision-aided, quaternion-based Iterated Kalman Filter (QEKF) \cite{QEKF} was developed to provide a general robotics solution for orientation estimation. In this work, we retain the orientation estimator that we developed in \cite{kang24}.
 % \TODO{add a subsection for legged state estimation}
 
\subsection{Generalized Momentum Dynamics}
The configuration of a legged robot is typically described using the generalized coordinates $q \in SE(3) \times \mathbb{R}^n$ that include the floating-base states:
\begin{equation}
    q = \begin{bmatrix}
        p^\intercal & [\phi \ \theta \ \psi]& \alpha^\intercal   
    \end{bmatrix} ^\intercal, 
\end{equation}
where $p$, $[\phi \ \theta \ \psi]$, and $\alpha \in \mathbb{R}^n$ represent the floating-base position, Euler angles in the world frame, and joint angles, respectively. $n$ denotes the dimension of the joint states. Consider the full-body dynamics of a legged robot:
\begin{equation}
     M(q) \Ddot{q} + C(q,\dot{q})\dot{q} + G(q) = B u + \textstyle{\sum}_i J_i(q)^T f_{i}, \textstyle \label{eq:full_dynamics}
\end{equation}
where $M(q)\in \mathbb{R}^{(6+n)\times (6+n)}$, $C(q,\dot{q}) \in \mathbb{R}^{(6+n)\times (6+n)}$ and $G(q)\in \mathbb{R}^{(6+n)}$ represent inertia matrix, Coriolis matrix, and gravitational vector, respectively. $B$ denotes a constant matrix that maps the joint torque vector $u$ to the generalized force vector. $\dot{q}$ and $\Ddot{q}$ denote the first and second-time derivatives of the generalized coordinates, and $f_i$ and $J_i(q)$ denote the contact GRF and its associated coordinate Jacobian, respectively. Specifically, for legged robots with point feet, the GRF $f_i$ acting on each individual foot $i$ is considered to have three degrees of freedom (DOF). 

Given the torques applied at the actuators $u$ and generalized coordinates accelerations $\Ddot{q}$, the expected trajectory outcomes based on the dynamics model can be compared to the actual results, with any discrepancies attributed to external wrenches. 
The momentum-based method leverages the properties of the Coriolis matrix, eliminating the need for noisy estimates or measurements of the second-order derivative $\Ddot{q}$. The generalized momentum $m$ and its derivatives $\dot{m}$ are defined respectively as:
\begin{equation}
   m =  M(q) \dot{q}, \quad \dot{m} = \dot{M(q)} \dot{q} + M(q) \Ddot{q}. 
\end{equation}
Based on the skew symmetry properties of $\dot{M(q)} - 2 C(q,\dot{q})$ and the symmetric property of $M(q)$, it can be derived that:
\begin{equation}
\dot{M(q)} = C(q,\dot{q}) + C(q,\dot{q})^T.    
\end{equation}
The dynamics of generalized momentum $m$ is expressed as:
\begin{equation}
    \dot{m} =   C(q,\dot{q})^T \dot{q} - G(q) + B u + \textstyle{\sum}_i J_i(q)^T f_{i}, \textstyle
\end{equation}
 which only relies on the estimation or measurements of $q$, $\dot{q}$, and $u$.

\section{Simultaneous Force and State Estimation}
This section introduces the proposed method for estimating GRF alongside the floating-base states of the robot. We formulate a decentralized Moving Horizon Estimation \cite{kang24}, incorporating additional momentum and contact constraints. The state \eqref{eq:state}, process model \eqref{eq:dynamic_constraints}, measurement model \eqref{eq:measurement_constraints}, and the physical constraints \eqref{eq:contact_constraints} in \eqref{equ:MHE} are detailed.

\subsection{Decentralized Estimation}
For general estimation tasks, a set of robot-centric states are selected to describe the motion of the legged robot, enabling simultaneous estimation of the robot's base position, velocity, and orientation. In the decentralized framework used in \cite{cheetah3}, \cite{HLIP}, the base orientation is estimated separately from the position and velocity, using a standalone estimator. Considering the unreliable performance of magnetometers in robotic applications and the terrain dependency of kinematics-based measurements, we proposed a vision-aided iterated QEKF in our previous work \cite{kang24}. This method iteratively fuses IMU readings with delayed VO corrections through synchronization and trajectory updates, consistently providing accurate orientation estimates. The estimation result $\hat{R}_\mathcal{WB}$ is used to simplify the following process and measurement models to linear time-varying (LTV) constraints of the MHE, as illustrated in Fig. \ref{Framework}. This decentralization transforms the MHE into a convex QP problem. The notation $\hat{(\cdot)}$ and $\tilde{(\cdot)}$ represent the estimated and measured quantity, respectively.     
\subsection{Process Model}
The state of the simultaneous force and state estimation is chosen to
describe the motion of the legged robot, the generalized momentum, and the ground reaction force:
\begin{equation}
    x = \begin{bmatrix}
        p^\intercal & v^\intercal & b_a^\intercal & m^\intercal & f_{i}^\intercal 
    \end{bmatrix}^\intercal ,
\end{equation}
where $p$, $v$, and $f$ denote the base position, base velocity, and GRF, respectively, in the world frame denoted by $\mathcal{W}$. $b_a$ denotes the accelerometer bias in the body frame $\mathcal{B}$. $m$ denotes the generalized momentum. For simplicity, only one of the $N$ legs is included in the notation. The actual number of feet considered depends on the specific robot and its locomotion behavior.
% Assuming relatively small effects of the floating base velocity $v$ on the Coriolis matrix, 

The state evaluation is modeled by the following time-varying linear discrete process model:
\begin{equation}
    x^{+} = \begin{bmatrix}
        p + v \Delta t + \frac{1}{2} \big(\hat{R}_{\mathcal{WB}} (\tilde{a} - b_{a}) + \text{g} \big) \Delta t^2\\
        v + \big(\hat{R}_{\mathcal{WB}} (\tilde{a} - b_{a}) + \text{g}\big) \Delta t\\
        b_{a}\\
        \textstyle m + \Delta t \dot{m} \\
        f_i
    \end{bmatrix} + \delta_x , \label{eq:linear_process} 
\end{equation}
where $\tilde{a}$ and $\text{g}$ denotes the accelerometer readings and the gravitational vector. $\delta^x \sim \mathcal{N}(0,Q)$ denotes the noise of the process model. The generalized momentum dynamics is linearized using time-varying Coriolis matrix:
\begin{equation}
    \dot{m} = C_1^\intercal v + C_{2}^\intercal \begin{bmatrix}
            \tilde{\omega}^\intercal & \tilde{\dot{\alpha}}^\intercal
        \end{bmatrix}^\intercal - G + B\tilde{u} + \textstyle{\sum}_i{J_{i}^\intercal f_{i}},
\end{equation}
where $C_1 \in \mathbb{R}^{3 \times (n+6)}$ and  $C_2 \in \mathbb{R}^{ (n+3) \times (n+6)}$ are sub-blocks of the Coriolis matrix $C = \begin{bmatrix}
    C_1^\intercal & C_2^\intercal
\end{bmatrix} ^ \intercal$, evaluated at the estimated generalized coordinates for the corresponding time. Considering the relatively small magnitude of the Coriolis matrix, the effects of linearization can be mitigated and treated as model uncertainties.

\subsection{Measurement Model}
\noindent{\textbf{Leg Odometry:}}
Once static contact is established between the foot and the ground, floating-base velocity $y_v$ is measured based on the known leg kinematics which relies on the encoder position and velocity measurements $\tilde{\alpha}$ and $\tilde{\dot{\alpha}}$: 
\begin{align}
    y_{v} &= v + \delta^y_{v}, \ ( \text{if} \ \tilde{b} = 1) , \label{eq:vfootmeas}\\
    \tilde{y}_{v} &=   - \hat{R}_{\mathcal{WB}} J_b(\tilde{\alpha}) \tilde{\dot{\alpha}} - \hat{R}_{\mathcal{WB}} \tilde{\omega}^{\times} fk(\tilde{\alpha}),
\end{align}
where $\tilde{b}$ is a boolean variable representing the contact state measurements of the foot, which is directly obtained from the contact sensor and $\tilde{\omega}$ is the angular velocity reading.  
The measurement noise $\delta^y_{v} \sim \mathcal{N}(0,R_v)$ combines multiple uncertainties, including the calibration and kinematics modeling error, encoder noises, and gyroscope noises. $J_b(\alpha)$ and $fk(\alpha)$ denote the body jacobian and forward kinematics, respectively.

\noindent{\textbf{Generalized Momentum:}}
Given the robot's dynamics and encoder measurements, the generalized momentum measurement model is formulated as:
% \begin{equation}
%     M = \begin{bmatrix}
%         M_1 & M_2
%     \end{bmatrix} ,
% \end{equation}
%  by similarly decoupling the inertial matrix $M = \begin{bmatrix}
%         M_1 & M_2
%     \end{bmatrix}$.
% where $M_1 \in \mathbb{R}^{(n+6)\times3}$ and  $M_2 \in \mathbb{R}^{(n+6)\times (n+3)}$. The generalized momentum measurement is then expressed as:
\begin{align}
    y_{m} &= m - M_1 v + \delta^y_{m} \label{eq:GM} ,\\
    \tilde{y}_{m} &= M_2 \begin{bmatrix}
        \tilde{\omega}^\intercal & \tilde{\dot{\alpha}}^\intercal 
    \end{bmatrix}^\intercal,
\end{align}
where $M_1 \in \mathbb{R}^{(n+6)\times3}$ and $M_2 \in \mathbb{R}^{(n+6)\times (n+3)}$ are subblock of inertia matrix $M = \begin{bmatrix}
        M_1 & M_2
    \end{bmatrix}$, and $\delta^y_m \sim \mathcal{N}(0,R_m)$ contains the measurement noise of $\omega$, $\alpha$ and $\dot{\alpha}$.

\noindent{\textbf{Visual Odometry:}}
Visual Odometry (VO) estimates the robot's pose in the camera frame $\mathcal{C}$ by tracking features in images captured by onboard cameras. Typically, the VO output is interpreted as the incremental homogeneous transformation between consecutive camera frames $\mathcal{C}_i$ and $\mathcal{C}_j$: 
\begin{align}
\mathbf{T}_{\mathcal{C}ij} = \mathbf{T}_{\mathcal{WC}_i}^{-1} \mathbf{T}_{\mathcal{WC}_j} + \delta_{\text{vo}}  ,  
\end{align}
where $\mathbf{T}_{\mathcal{WC}_i}$ and $\mathbf{T}_{\mathcal{WC}_j}$ are the homogeneous transformations from the world frame $\mathcal{W}$ to the camera frame $\mathcal{C}$ at time $i$ and $j$. The VO measurement is corrupted by noise $\delta_{\text{vo}} \sim \mathcal{N}(0,Q_{\text{vo}})$. With the transformation $\mathbf{T}_{\mathcal{BC}}$ between IMU/body frame $\mathcal{B}$ and camera frame $\mathcal{C}$ calibrated offline, the incremental transformation $\mathbf{T}_{\mathcal{C}_{ij}}$ can be represented in the body frame as:
\begin{equation}
    \tilde{\mathbf{T}}_{\mathcal{B}_{ij}} = \mathbf{T}_{\mathcal{BC}} \tilde{\mathbf{T}}_{\mathcal{C}_{ij}} \mathbf{T}_{\mathcal{BC}}^{-1} . \label{eq:VO_body}
\end{equation}
The output frequency of the VO matches the camera frame rate, ranging from 10 to 60 Hz. 
In our implementation, each camera/VO frame is aligned to the closest IMU frame on the time axis. After the alignment, only the translation component $p_{\mathcal{B}_{ij}}$ of the incremental transformation $\mathbf{T}_{\mathcal{B}_{ij}}$ is applied as a measurement into the estimation.
We employ a Cubic Bézier polynomial to fit a series of relative translation measurements $\tilde{p}_{\mathcal{B}_{ij}}$. A smooth path of position is generated using consecutive translations $\tilde{p}_{\mathcal{B}_{ij}}$ as control points, providing interpolated VO measurements $\tilde{p}_{c_{k}}$ between consecutive IMU frames $k$ and $k+1$, $\forall k \in \{k \in \mathbb{N}^+ | \ i \leq k \leq j-1\}$:
\begin{equation}
    y_{c_k} = \hat{R}_{\mathcal{WB}}^\intercal (p_{k+1} - p_{k}) + \delta^y_{c_k}, \quad \tilde{y}_{c_k} = \tilde{p}_{c_k} , \label{eq:cam_measure} \\
\end{equation}
where $\tilde{p}_{c_k}$ is the interpolated transformation between IMU frames. $\delta^y_c \sim \mathcal{N}(0,R_c)$ denotes the VO measurements noise. 

\subsection{Physical Constraints}\
In \cite{lcpconstrainedKF}, the Linear Complementary Problem (LCP) is formulated with the maximum dissipation principle and a polyhedral friction cone, leading to the development of a constrained Kalman Filter for handling rigid slippery contacts. In this work, we employ a simplified LCP to integrate physical information as constraints in the estimation. For static rigid contact, the complementarity constraint is formulated as:
% \begin{align}
%     v_\text{foot} \tilde{b} &= 0, \quad \label{eq:guard}\\
%     f (1 - \tilde{b}) &= 0, \quad f_\text{normal} \geq 0 \label{eq:comp1}
% \end{align}
\begin{align}
    v_\text{foot} \tilde{b} &= 0, \quad f (1 - \tilde{b}) = 0, \quad f_\text{normal} \geq 0, \label{eq:guard}
\end{align}
where $v_\text{foot}$ and $f_\text{normal}$ denote the foot velocity and the normal component of the linear part of GRF acting on the corresponding robot foot, respectively. This simplification leverages the advantage of using a pressure sensor or contact switch as contact indicator $\tilde{b}$, addressing the computational challenges introduced by the original mixed-integer formulation \cite{lcpconstrainedKF}. When walking on slippery terrain, the velocity constraint of the contact foot reduces to $v_\text{foot}^n \tilde{b} = 0$.  

\subsection{Linear Moving Horizon Estimation}
With linear constraints constructed above, the \eqref{equ:MHE} is formulated as a constrained Quadratic Program:
\begin{align}
&\min_{\mathbf{X}_{[T-N,T]}} \  \Gamma(x_{T-N}) \textstyle \nonumber \\
&+ \textstyle{\sum}_{k = T-N}^{T-1} || \delta^x_{k} ||_{Q_k} + \textstyle{\sum}_{k=T-N}^T || \delta^y_k ||_{R_k} , \\
&\text{s.t.} \quad \eqref{eq:linear_process},  \eqref{eq:vfootmeas}, \eqref{eq:GM}, \eqref{eq:cam_measure}, \eqref{eq:guard} \nonumber .
\end{align}
The computation burden of canonical MHE is addressed with this convex optimization formulation. With a proper selection of arrival cost term $\Gamma(x_{T-N})$, the MHE iteratively solves the MAP problem, incorporating history sensor information from outside the current window.
% \section{Orientation Estimation}
%  For nine-axis IMU, off-the-shelf AHRS (Attitude and Heading Reference Systems) are able to provide a reasonable orientation measurement $\hat{R}_{\text{WB}}$ relative to the direction of gravity and the earth magnetic field \cite{madgwick2010efficient}. 

% \section{Arrival Cost Calculation}
\begin{figure}[H]
    \centering
    \setlength{\abovecaptionskip}{0pt}
            \vspace{-5pt}
    \includegraphics[width=1 \linewidth]{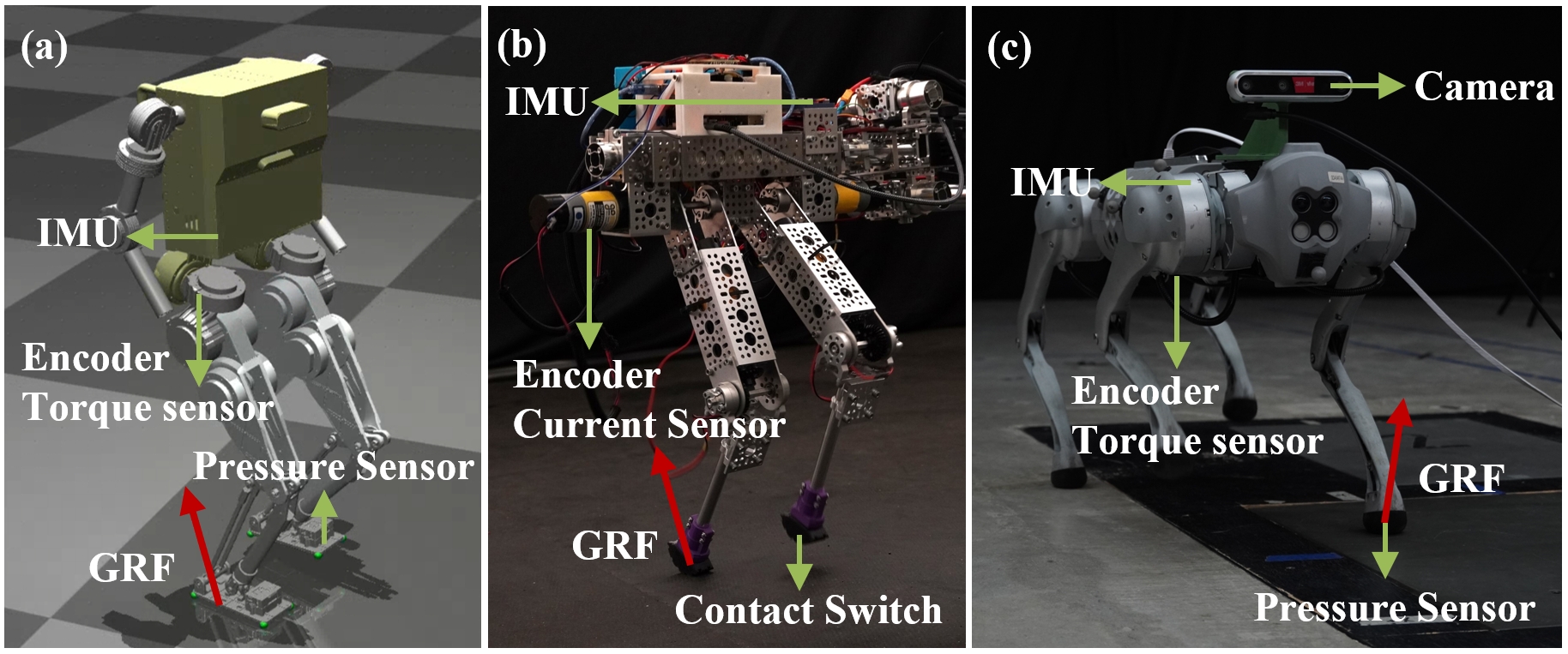}
    \vspace{-10pt}
    \caption{Sensor configurations of (a) Bucky and (b) STRIDE. (c) Sensor configuration and experimental setup of the Unitree Go1.}
    \label{experiment_illu}
    \vspace{-15pt}
\end{figure}
\begin{figure}[H]
    \centering
    \setlength{\abovecaptionskip}{0pt}
    \includegraphics[width=.99\linewidth]{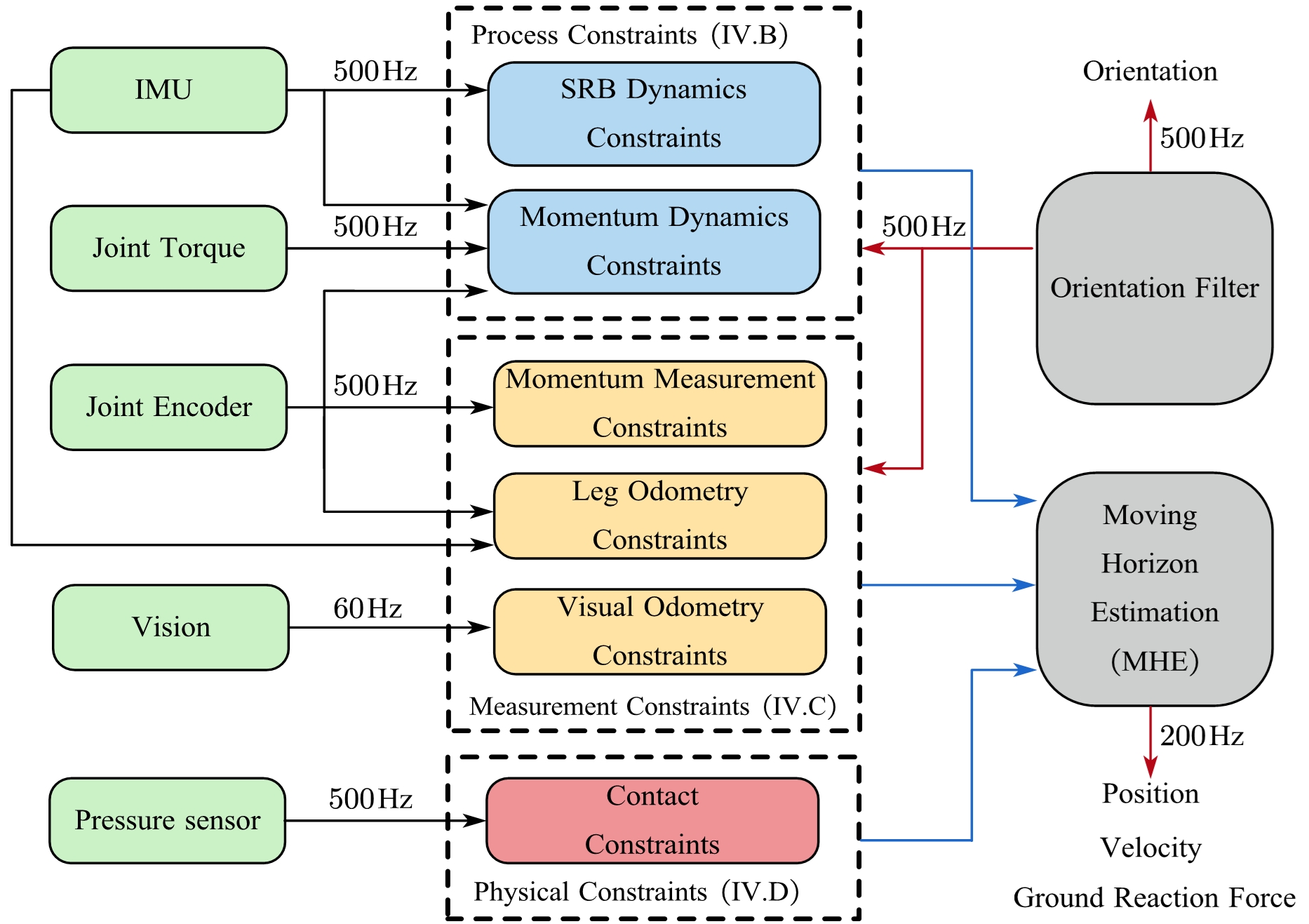}
    \vspace{-10pt}
    \caption{The proposed decentralized estimation framework with its application to the quadrupedal robot Go1.}
    \label{framework}
    % \vspace{-5pt}
\end{figure}
\section{Evaluation}
Now we evaluate the proposed simultaneous force and state estimation method in both simulation and hardware, using the custom-designed humanoid robot Bucky, the educational planar bipedal robot STRIDE~\cite{Stride} and commercial quadrupedal robot Unitree Go1, as illustrated in Fig. \ref{experiment_illu}. The MHE is implemented using C++ in ROS2 and Ubuntu 20.04. The QP problem is solved using the off-the-shelf solver OSQP~\cite{stellato2020osqp}. The VO is implemented through the open-source package ORB-SLAM3~\cite{Campos_2021} to perform VO at 60Hz. The MHEs are solved stably at 200Hz, with a window size of 8 that contains 1-2 VO frames in practice. All state estimations are performed in real-time on a PC with Intel Core i9-13900 CPU. Additionally, the STRIDE and Bucky robots are controlled using the S2S controller \cite{HLIP}, \cite{xiong2019orbit}, and the Go1 robot is controlled using Model Predictive Control (MPC) and Whole body control (WBC) framework \cite{MPC}, \cite{WBC}, \cite{liao2023walking}.
\begin{table}[H]
\centering
\caption{Sensor Noise of Bucky, STRIDE and Go1 in simulation.}
        \vspace{-5pt}
\begin{tabular}{|c|c|}
\hline
Sensor Type & Noise std.\\
\hline
Accelerometer & 0.04 [m/s$^2$]\\
\hline
Gyroscope &   0.002 [rad/s]\\
\hline
Encoder velocity &     0.02 [rad/s]\\
\hline
Encoder position &     0.01 [rad] \\
\hline
Actuator effort  &    0.01 [N $\cdot$ m] \\ 
\hline
\end{tabular}
\label{table:noise}
        \vspace{-10pt}
\end{table}

\begin{table}[H]
    \centering
    \caption{RMSEs of the estimations on STRIDE in simulation.}
     \vspace{-5pt} % don't use it too aggressively
    \begin{tabular}{|>{\centering\arraybackslash}p{1.9cm}|>{\centering\arraybackslash}p{1.4cm}|>{\centering\arraybackslash}p{1.4cm}|>{\centering\arraybackslash}c|>{\centering\arraybackslash}c|}
    \hline
    Estimation & \multirow{2}{*}{Our Method} & Our Method & \multirow{2}{*}{DKF} & \multirow{2}{*}{MBO} \\
    Method & & (w/o IV.D) & & \\
    \hline
    $\text{RMSE}_v$ [m/s]  &    0.0296   &  0.0305   &    0.0285 &  NA \\
    \hline
    $\text{RMSE}_f$ [N]&     1.7617   &  2.4277 &  2.8613  &   2.7460 \\
    \hline
    \end{tabular}
    \label{table:stride}
            \vspace{-5pt}
\end{table}

\begin{figure}[H]
    \centering
    \setlength{\abovecaptionskip}{0pt}
    % \includesvg[width=1.0\linewidth]{figure/stride_exp.svg}
    \includegraphics[width=1.0\linewidth]{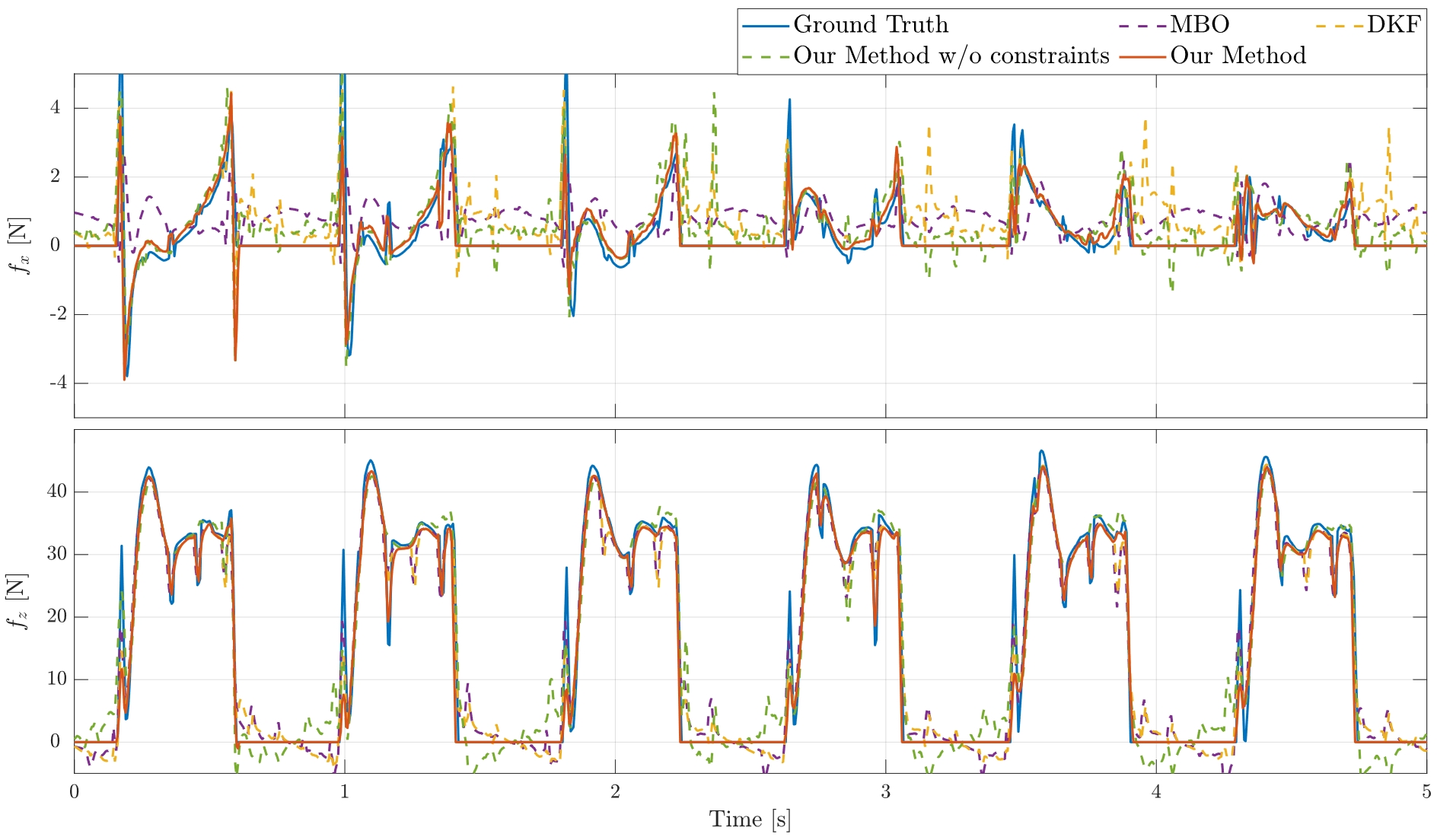}
        \vspace{-12pt}
    \caption{Estimation results on STRIDE in simulation: $f_z$ and $f_x$ denote the vertical normal force and the tangential frictional force, respectively.}
        \vspace{-10pt}
    \label{stride_sim}
\end{figure}

\begin{table}[t]
    \centering
    \caption{RMSEs of the estimations on Bucky in simulation.}
     \vspace{-5pt} % don't use it too aggressively
    \begin{tabular}{|>{\centering\arraybackslash}p{1.9cm}|>{\centering\arraybackslash}p{1.4cm}|>{\centering\arraybackslash}p{1.4cm}|>{\centering\arraybackslash}c|>{\centering\arraybackslash}c|}
    \hline
    Estimation & \multirow{2}{*}{Our Method} & Our Method & \multirow{2}{*}{DKF} & \multirow{2}{*}{MBO} \\
    Method & & (w/o IV.D) & & \\
    \hline
    $\text{RMSE}_v$ [m/s]  &    0.0161   &  0.0165   &    0.0153 &  NA \\
    \hline
    $\text{RMSE}_f$ [N]&       10.4337   &  24.6590 &  33.1068  &   10.1655 \\
    \hline
    \end{tabular}
    \label{table:bucky}
            \vspace{-5pt}
\end{table}

\begin{figure}[H]
    \centering
    \vspace{-2pt}
    \setlength{\abovecaptionskip}{0pt}
    % \includesvg[width=1.15\linewidth]{figure/bucky_sim.svg}
    \includegraphics[width=1\linewidth]{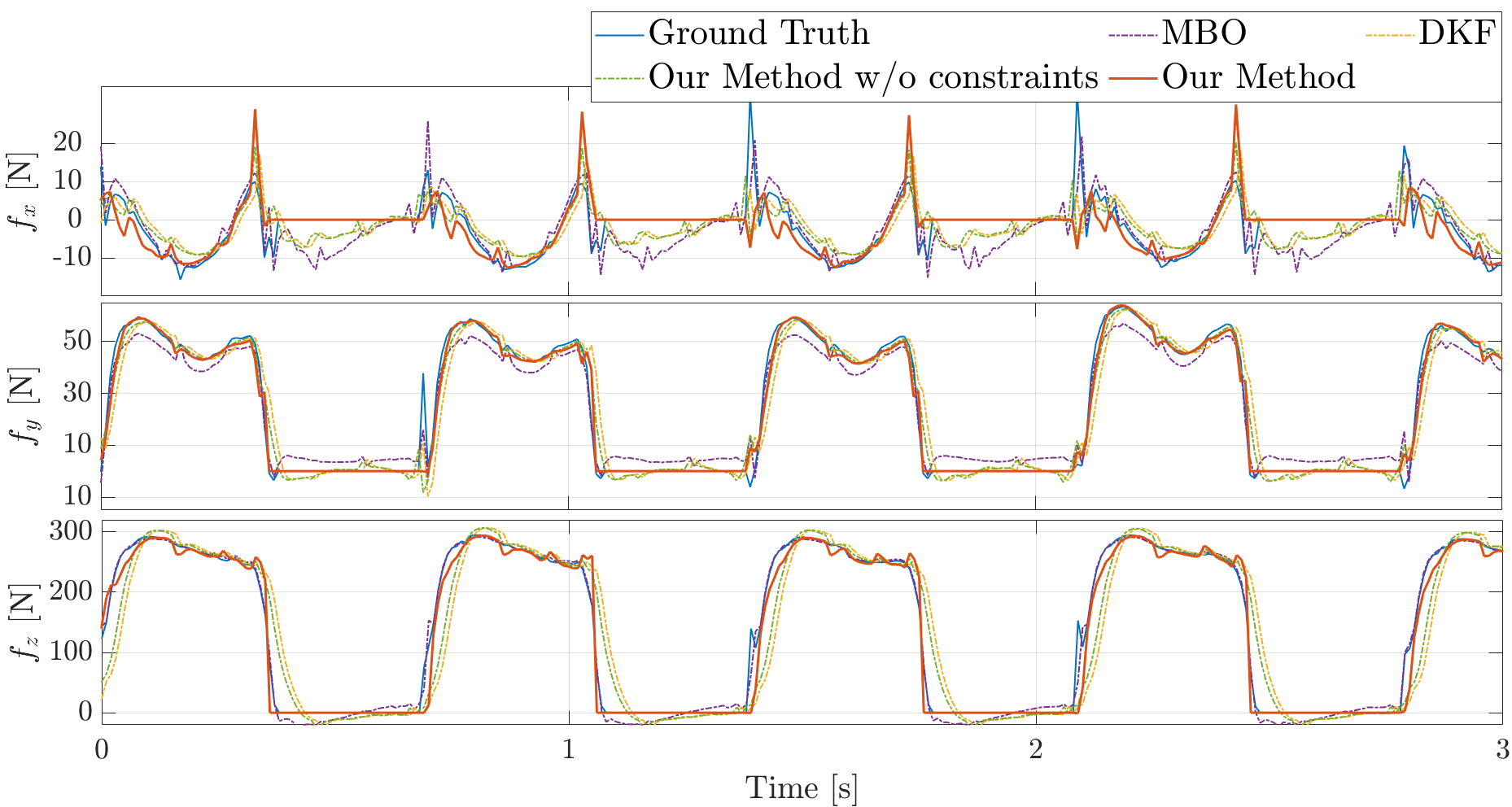}
        \vspace{-10pt}
    \caption{Estimation results of the GRF on Bucky in simulation, $f_x$, $f_y$, and $f_z$ denote the tangential frictional force and the vertical normal force, respectively, expressed in the world frame.}
        \vspace{-5pt}
    \label{bucky_sim}
\end{figure}

The simulation is configured with the MuJoCo \cite{mujoco}, which runs the time-stepping simulation at 500 Hz. The sensor noise settings for the MuJoCo environment are detailed in Table. \ref{table:noise}, and the ground truth GRF is obtained from a force sensor attached to the foot. For hardware experiments, we use a motion-capture system that has 12 OptiTrack Prime-13 cameras \cite{Optitrackebsite} to obtain the ground truth position and orientation at 200Hz. The ground truth velocity is obtained by computing the central differences of the position. The GRF ground truth is obtained from a Bertec-4060 6D Force Plate \cite{forceplatewebsite} at 1000Hz, with a resolution of $\pm 0.4\text{N}$. Fig. \ref{framework} illustrates the system architecture for estimation.

We focus on comparing the GRF estimation errors of a modified Disturbance Kalman Filter (DKF) \cite{DisturbKF}, Momentum-Based Observer (MBO) \cite{View}, the proposed MHE method without physical constraints described in Section IV.D, and the full proposed MHE method. The modified DKF is formulated to be equivalent to an unconstrained MHE with a window size of one. For the MBO, we use the ground truth floating-base velocity to construct the generalized momentum, creating an idealized scenario for comparison.

\subsection{Humanoid Bucky and Bipedal STRIDE Simulations}

We evaluate our estimation method using simulation data from the humanoid robot Bucky and planar bipedal robot STRIDE \cite{Stride}. Fig. \ref{stride_sim}, Fig. \ref{bucky_sim}, Table. \ref{table:stride}, and Table. \ref{table:bucky} show our methods closely match the ground truth, demonstrating that the proposed approach outperforms the MBO and DKF methods. Additionally, the inclusion of contact complementarity constraints, which enforces known physical information, further enhances the accuracy of the estimation.

\subsection{Unitree Go1 Simulation}
\begin{figure}[t]
    \centering
    \setlength{\abovecaptionskip}{0pt}
    % \includesvg[width=1.0\linewidth]{figure/go1_sim.svg}
    \includegraphics[width=1.0\linewidth]{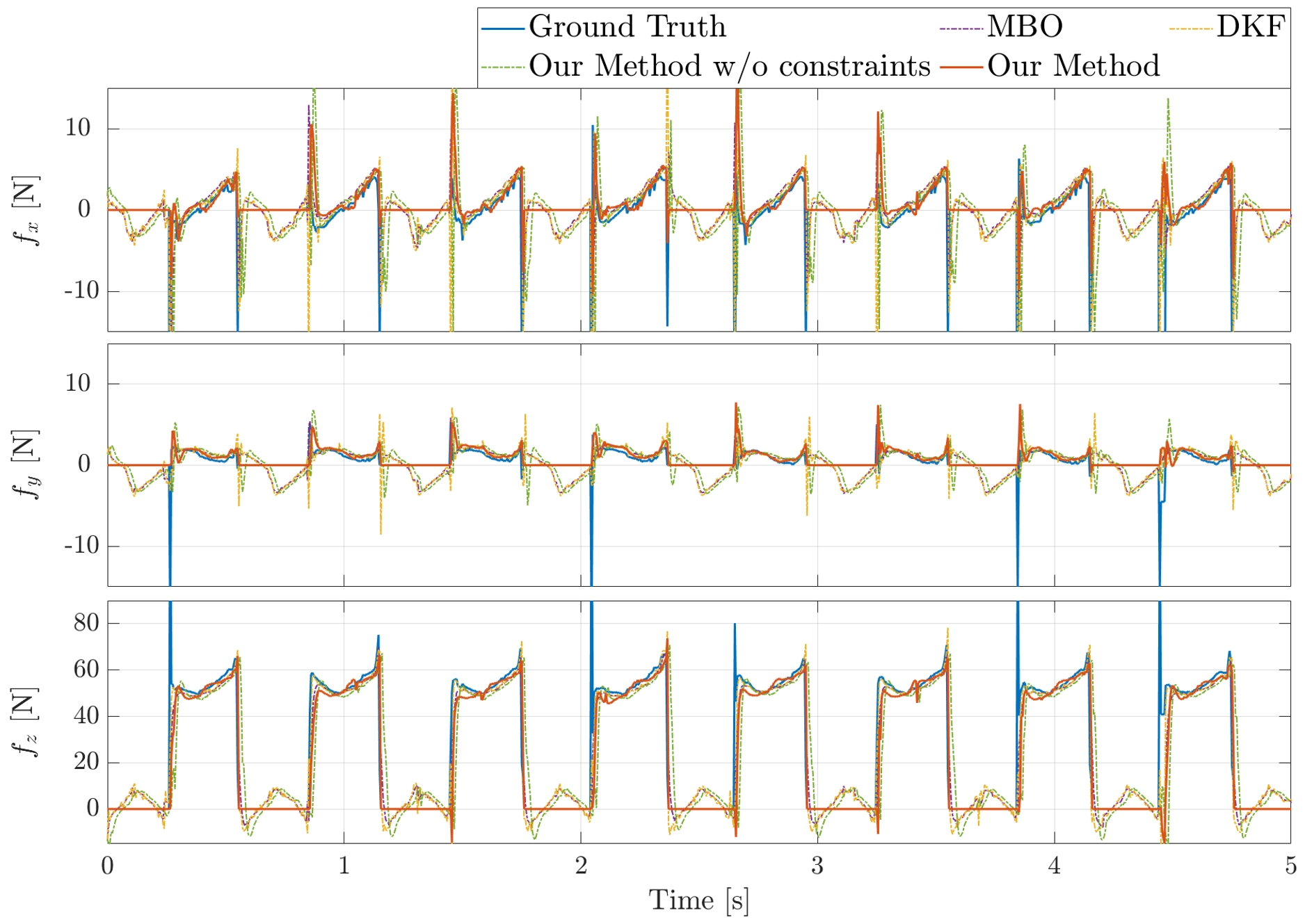}
    \vspace{-12pt}
    \caption{Estimation results of the GRF on the Unitree Go1 in simulation.}
    \label{go1_sim}
    \vspace{-15pt}
\end{figure}

We evaluate our method using simulation data from the Go1 quadruped robot. In simulation, the Go1 robot is controlled with a basic velocity command to execute omnidirectional locomotion with various gaits. Fig. \ref{go1_sim} and Table. \ref{table:go1_sim} show the estimation results of floating-base velocity and GRF on the front left foot. The proposed method demonstrates superior performance compared to canonical approaches.

\begin{table}[t]
    \centering
    \caption{RMSEs of the estimations on Go1 in simulation.}
     \vspace{-5pt} % don't use it too aggressively
    \begin{tabular}{|>{\centering\arraybackslash}p{1.9cm}|>{\centering\arraybackslash}p{1.4cm}|>{\centering\arraybackslash}p{1.4cm}|>{\centering\arraybackslash}c|>{\centering\arraybackslash}c|}
    \hline
    Estimation & \multirow{2}{*}{Our Method} & Our Method & \multirow{2}{*}{DKF} & \multirow{2}{*}{MBO} \\
    Method & & (w/o IV.D) & & \\
    \hline
    $\text{RMSE}_v$ [m/s]  &  0.0146  &  0.0148   &    0.0133 &  NA \\
    \hline
    $\text{RMSE}_f$ [N]&  3.6176 &  6.9333 &  5.1350  &    4.8948 \\
    \hline
    \end{tabular}
        % \vspace{-5pt}
    \label{table:go1_sim}
\end{table} 

\subsection{Unitree Go1 Hardware}

\begin{table}[t]
    \centering
         % \vspace{-5pt} % don't use it too aggressively
    \caption{RMSEs of the estimations on Go1 hardware.}
     \vspace{-5pt} % don't use it too aggressively
    \begin{tabular}{|>{\centering\arraybackslash}p{1.9cm}|>{\centering\arraybackslash}p{1.4cm}|>{\centering\arraybackslash}p{1.4cm}|>{\centering\arraybackslash}c|>{\centering\arraybackslash}c|}
    \hline
    Estimation & \multirow{2}{*}{Our Method} & Our Method & \multirow{2}{*}{DKF} & \multirow{2}{*}{MBO} \\
    Method & & (w/o IV.D) & & \\
    \hline
    $\text{RMSE}_v$ [m/s]  &  0.0504 &  0.0497   &   0.0628
 &  NA \\
    \hline
    $\text{RMSE}_f$ [N]&  6.6790  &  11.2255 & 10.1594  &    12.7229
 \\
    \hline
    \end{tabular}
    \label{table:go1_hard}
            \vspace{-5pt}
\end{table} 

\begin{figure}[t]
    \centering
    \setlength{\abovecaptionskip}{0pt}
    % \includesvg[width=1.0\linewidth]{figure/go1_real_add.svg}
    \includegraphics[width=1.0\linewidth]{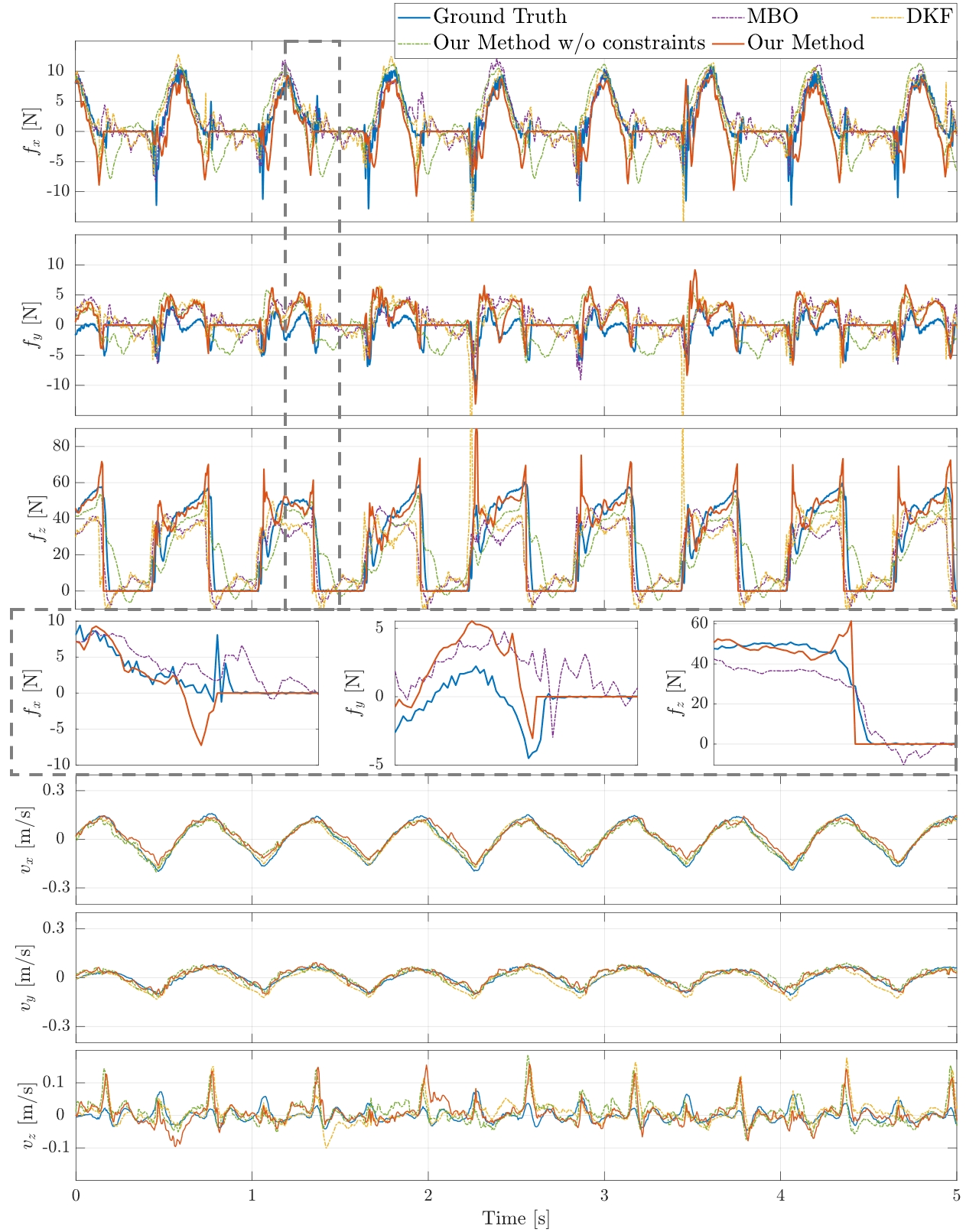}
    \vspace{-15pt}
    % \caption{Estimation results of the GRF and linear velocity on the Unitree Go1 in hardware.}
    \caption{Estimation results of GRF and linear velocity on the Unitree Go1 hardware: GRF is expressed in the world frame and linear velocity in the body frame. The grey-dotted box is magnified to compare our method with the ground truth and the trending MBO method.}    
    \label{go1_force_hard}
            \vspace{-15pt}
\end{figure}

% \begin{figure}[t]
%     \centering
%     \setlength{\abovecaptionskip}{0pt}
%     \includesvg[width=1.0 \linewidth]{figure/go1_vel.svg}
%     \caption{Estimation results of linear velocity on Unitree Go1 in hardware.}
%     \label{go1_vel_hard}
% \end{figure}

We finally demonstrate our estimation method on the Go1 hardware. We obtain the stereo images from the Intel Realsense Depth Camera D455 and IMU data from the default Unitree IMU. During the experiments, the Go1 robot is controlled to trot on the force plate, as shown in Fig. \ref{experiment_illu}. The comparison results between the estimation and ground truth of the GRF on the front left foot and base-linear velocity are shown in Fig. \ref{go1_force_hard}. As indicated in Table. \ref{table:go1_hard}, for the noisy hardware implementation, the constraint information effectively enforces the feasibility of the estimation results, leading to superior performance, even though the momentum-based methods are configured to use the ground truth floating-base velocity from the opti-track system. The inclusion of exteroceptive sensors improves the estimation during the stance phase, while the contact constraints further ensure the feasibility of the solution.

\section{Conclusions and Future Work}
This paper presents a simultaneous ground reaction force and state estimation framework for legged robots, based on decentralized Moving Horizon Estimation (MHE). A convex windowed optimization is formulated to address sensor noise and the coupling between states and dynamics, resulting in a modular approach that incorporates generalized momentum dynamics, proprioceptive sensors, exteroceptive sensors, and contact complementarity constraints. Experiments conducted on various legged systems demonstrate the superior performance of the proposed methods compared to traditional momentum-based approaches.

In the future, we aim to explore terrain dynamics by leveraging estimated foot position, velocity, and GRF to perform planning and control tasks on deformable and unstructured terrains. We will also focus on designing computational algorithms to improve estimation frequency and window size while optimizing the calibration of process and measurement noises. Additionally, we are interested in integrating the Linear Complementarity Problem (LCP) within the windowed estimation framework to enable robust estimation in environments with rigid and slippery contact conditions.

%%%%%%%%%%%%%%%%%%%%%%%%%%%%%%%%%%%%%%%%%%%%%%%%%%%%%%%%%%%%%%%%%%%%%%%%%%%%%%%%

%%%%%%%%%%%%%%%%%%%%%%%%%%%%%%%%%%%%%%%%%%%%%%%%%%%%%%%%%%%%%%%%%%%%%%%%%%%%%%%%

%%%%%%%%%%%%%%%%%%%%%%%%%%%%%%%%%%%%%%%%%%%%%%%%%%%%%%%%%%%%%%%%%%%%%%%%%%%%%%%%

%%%%%%%%%%%%%%%%%%%%%%%%%%%%%%%%%%%%%%%%%%%%%%%%%%%%%%%%%%%%%%%%%%%%%%%%%%%%%%%%
\newpage
\bibliographystyle{IEEEtran}
\bibliography{reference}
% \addtolength{\textheight}{-5.0cm}

\end{document}